\documentclass[prl,floatfix,twocolumn,preprintnumbers,amsmath,amssymb,superscriptaddress]{revtex4}
\usepackage{color}

\usepackage{graphicx,psfrag}
\usepackage{hyperref}
\usepackage{verbatim}
\usepackage[utf8]{inputenc}
\usepackage{float}
\usepackage{sidecap}

\begin{document}
\title{GANs for generating EFT models}
\author{Harold Erbin}
\author{Sven Krippendorf}
\affiliation{{\it \small Arnold Sommerfeld Center for Theoretical Physics, LMU, Theresienstr.~37, 80333 M\"unchen, Germany}}
\preprint{LMU-ASC 58/18}
\begin{abstract}

We initiate a way of generating models by the computer, satisfying both experimental and theoretical constraints. In particular, we present a framework which allows the generation of effective field theories. We use Generative Adversarial Networks to generate these models and we generate examples which go beyond the examples known to the machine.
As a starting point, we apply this idea to the generation of supersymmetric field theories. In this case, the machine knows consistent examples of supersymmetric field theories with a single field and generates new examples of such theories. In the generated potentials we find distinct properties, here the number of minima in the scalar potential, with values not found in the training data. We comment on potential further applications of this framework.

\end{abstract}
\maketitle

A key activity in fundamental physics is to come up with models satisfying experimental constraints and theoretical paradigms. Finding such solutions requires human experience, imagination, and intuition on which extensions to consider. Only very rare situations allow a complete classification of solutions and human exploration generically is limited in time and sometimes imagination. It would be very exciting to explore automated model generation and to see which model building potential, machines can have in the context of fundamental physics. The aim is to have a tool which can generate models with pre-defined properties. Our effort comes at a time when machines are able -- though in different settings -- to come up with `creative' solutions to problems going beyond human capability such as in the context of AlphaGoZero~\cite{silver2017mastering}.

The language, in which models are formulated in fundamental physics, is that of effective field theories, which can, in many cases, be characterised by a field theory Lagrangian. The latter determines the couplings among fields and their respective dynamics. Theoretical and experimental constraints are implemented by structure in the couplings, for instance by requiring invariance of the Lagrangian under symmetries, such as invariance under the Lorentz group. Given a particular requirement, such as invariance under spacetime symmetries, it is a common problem to find theories consistent with such a symmetry. Such a list of requirements can be seen as the rules of the game which are imposed on the allowed models. The goal is to explore the space of models which are consistent with these symmetries and to determine which type of dynamics can appear. In most cases, physicists know consistent examples but do not know the general space of solutions. Finding consistent solutions which go beyond the known types of solutions is a common problem in physics. 

As a first example in this direction, we automatise the search for new supersymmetric models. Supersymmetry (SUSY) is one of the leading candidates for Beyond-The-Standard-Model physics (BSM), potentially addressing the electroweak hierarchy problem and is preferred in ultraviolet complete theories arising in string theory. Large experimental efforts are taken to search for low-energy remnants of supersymmetry at colliders and as dark matter candidates. The low-energy observables of supersymmetry crucially depend on how supersymmetry is broken. In the absence of gravity, i.e.~in the global limit of supersymmetry, the models of supersymmetry breaking are relatively limited, two prominent classes being~\cite{ORaifeartaigh:1975nky,Intriligator:2006dd}. An extension of the available models of supersymmetry breaking, potentially leading to different phenomenological signatures, is still highly desirable.

From a theoretical point of view, supersymmetric models allow a very tractable avenue for automatisation strategies. The simplest setup is that of a single chiral superfield with no gauge symmetries. In this context, taking canonical kinetic terms, the superpotential governs all dynamics. This superpotential is a holomorphic function in one variable. To generate new models becomes the task of generating holomorphic functions. Additional properties, such as the number of minima of the scalar potential or the masses in the minimum could be added as further requirements. 

As a first step, we restrict ourselves in this paper to generating superpotentials for a single field. Put concretely, we build a generator for a single field superpotential in a box, which is discretised. The problem of generating such a superpotential is then equivalent to generating an image with two colour channels~\footnote{An image is represented by a tensor with dimensions given by the width, height and number of colour channels. A black and white image has one  channel, whereas an RGB image has three channels.} and a particular local property, holomorphicity. Holomorphicity can be checked locally whether the Cauchy-Riemann equations for a function $f(z=x+iy)=u(x,y)+i v(x,y)$ are satisfied:
\begin{equation}
 \frac{\partial u(x,y)}{\partial x}=\frac{\partial v(x,y)}{\partial y}~,~\frac{\partial u(x,y)}{\partial y}=-\frac{\partial v(x,y)}{\partial x}~.
 \label{eq:CRholomorph}
\end{equation}
A function is holomorphic when these conditions are satisfied everywhere.

In this paper, we present numerical examples based on a $64\times 64$ grid, allowing for simulations to be carried out on `standard' GPUs of a desktop in reasonable time. The output has two channels, implementing the fact that we are interested in a complex-valued function. To build such a generator we use a Generative Adversarial Network (GAN) structure~\cite{goodfellow}. Such GANs have been extremely successful in generating images with particular properties \footnote{This is a very active field of research of which activities we are only partially aware. Probably, further efforts in network design will lead to a significant improvement in the generation of interesting superpotentials.}. As part of applications of machine learning in particle and astrophysics (cf.~\cite{Albertsson:2018maf,darkmachines}), GANs are also particularly useful in physics when circumventing costly simulations such as detector simulations~\cite{Paganini:2017hrr} or galaxy shape measurements for dark energy surveys~\cite{Ravanbakhsh:2016xpe}.  

We would like to stress that one difficulty of GANs is the reconstruction of global features (e.g.~generating images of animals with the correct numbers of characteristic features, examples can be found in~\cite{distorted}). In our case, this is not a `bug' but a feature. Globally distinct, but locally inseparable features are actually particularly interesting in the context of superpotentials. Here, this can correspond for instance to multiple minima of the scalar potential, which is highly relevant in models of early Universe cosmology. This seems to be a very intriguing avenue for model building, which the machine is performing here as it is combining a lot of known local features to a new global structure. This is precisely what is done in a lot of BSM model building, e.g.~in bottom-up string model building~\cite{Aldazabal:2000sa}.\\

\begin{figure}
\includegraphics[width=0.4\textwidth]{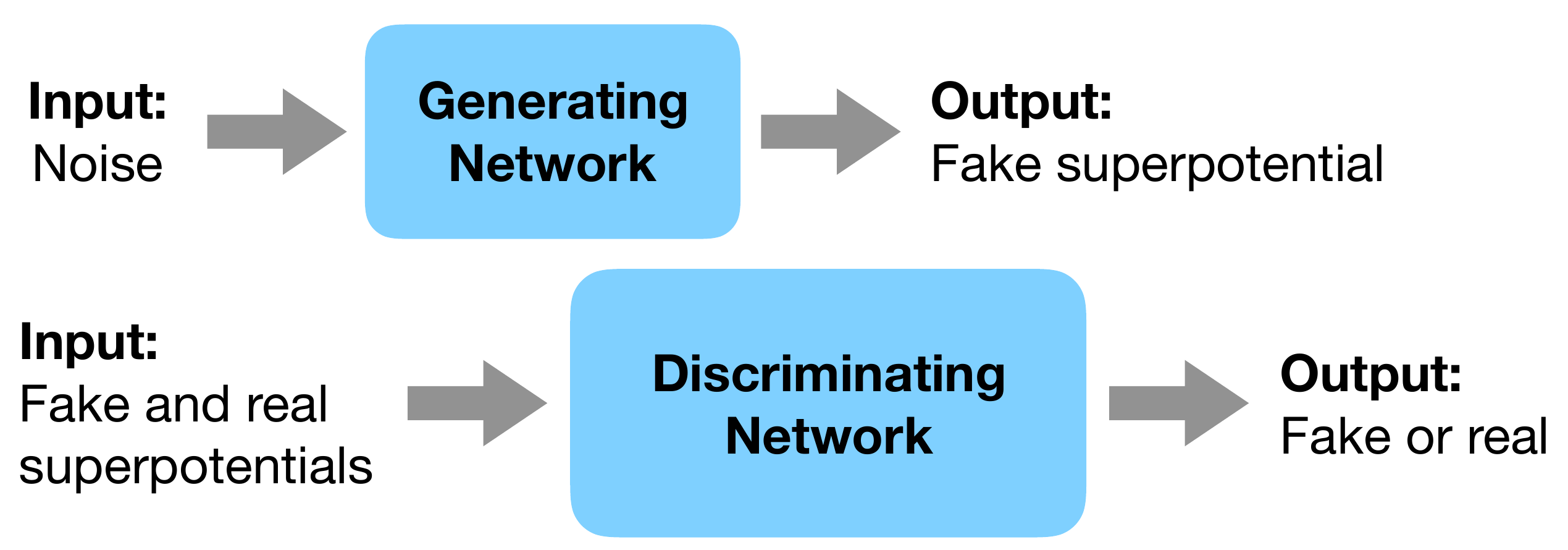}
\hspace{0.5cm} \includegraphics[width=0.45\textwidth]{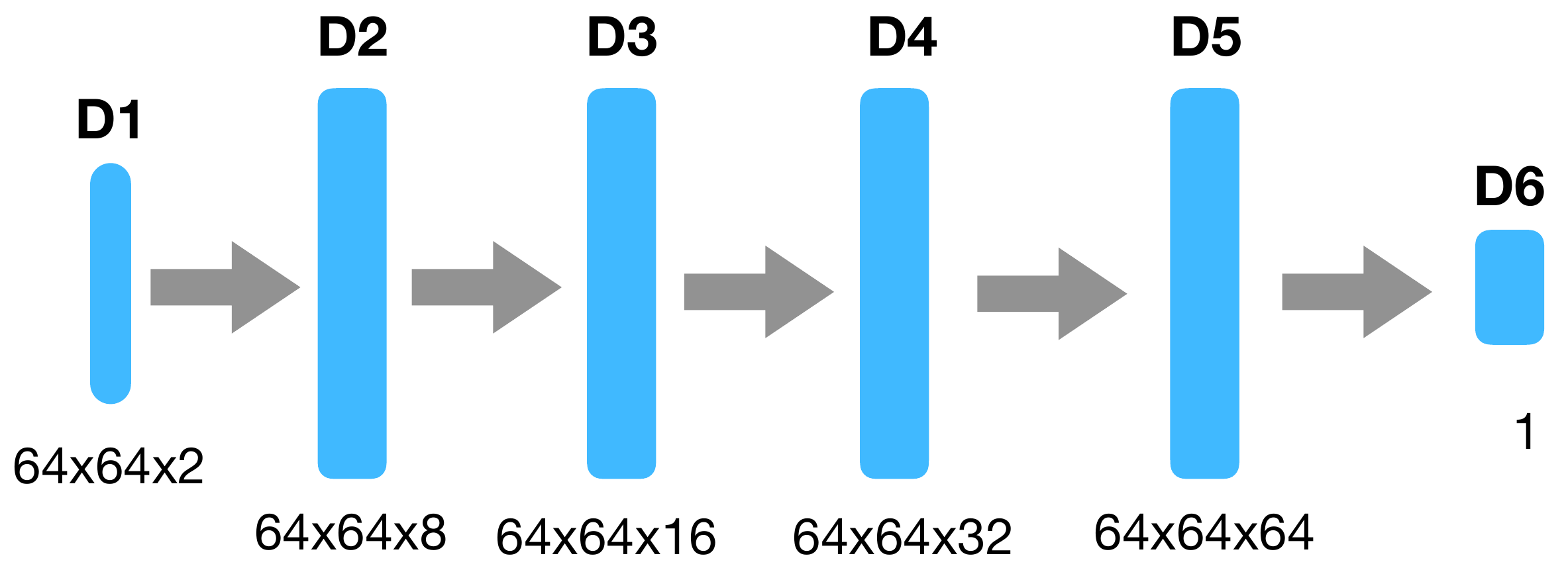}\\
\flushright \includegraphics[width=0.45\textwidth]{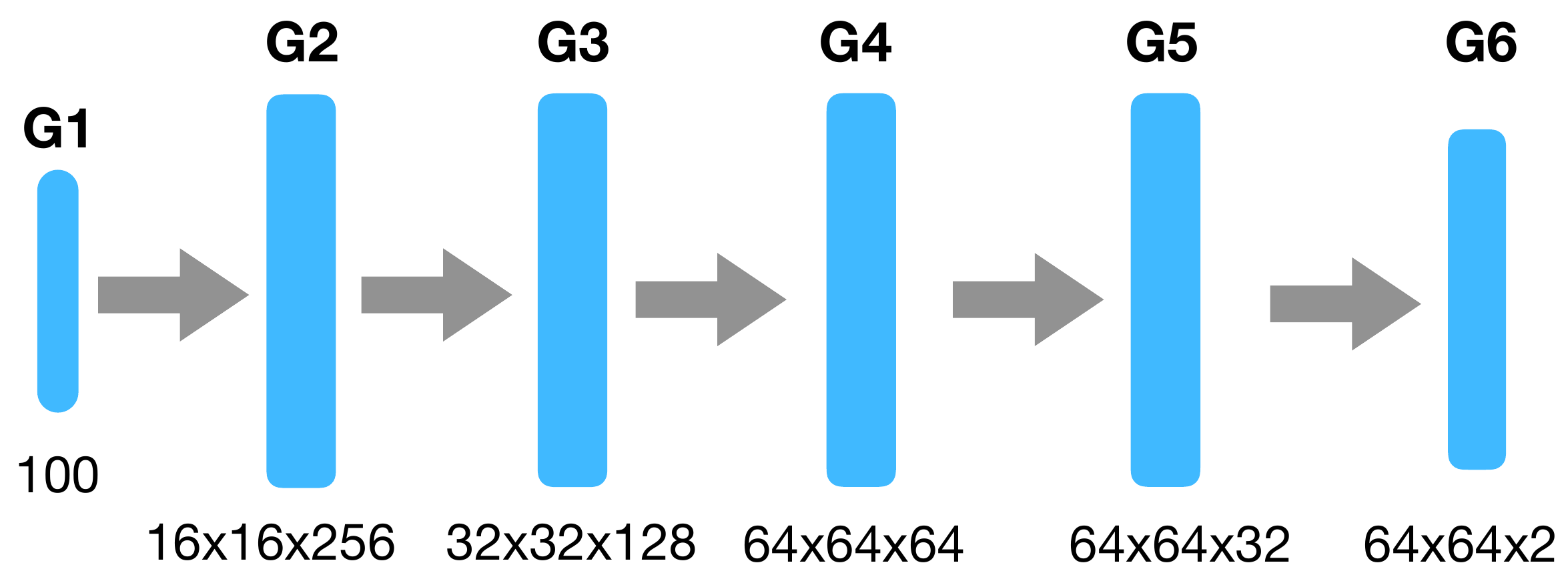}
\caption{Top: Overall design of the GAN framework. Middle: The layout of the discriminator network. The numbers indicate the respective output dimensions of the block of layers. D1 is the input layer. D2-D5 is a combination of a convolution layer, a LeakyReLU activation, and a dropout layer. D6 is a dense layer with a sigmoid activation. Bottom: The layout of the generator network. G1 is the noise input layer. G2 consists of a dense layer, batch normalisation, linear activation, and a dropout layer. G3-G4 consist of an upsampling layer followed by a convolutional layer, batch normalisation, and linear activation. G5 does not contain an upsampling layer but the same type of layers as G3-G4. G6 is a convolutional layer and a $tanh$ activation. A table with the exact layer structure for both the discriminator and the generator can be found at the end of this article.\label{fig:overalldesign}}
\end{figure}

\noindent{\bf Numerical setup:}\\
The basic idea of GANs is that two networks, the discriminating and generating network, are trained to compete against each other: the discriminating network is optimised to distinguish between real and fake data, whereas the generating network is optimised to produce fake data which tricks the discriminating network. In our case, the input for the discriminator network consists of generated images from the generating network and examples of superpotentials which we have generated from some known holomorphic functions. The overall structure of the network is shown in Figure~\ref{fig:overalldesign} at the top. 

\begin{SCfigure}
\includegraphics[width=0.3\textwidth]{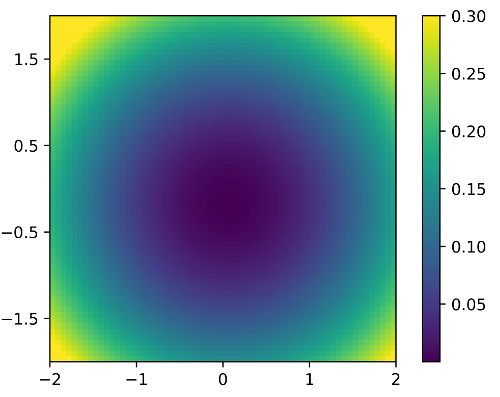}
\vspace{-0.5cm}
\caption{Example of a scalar potential associated to a polynomial superpotential from our training set. The initial range of coefficients is $\{-1,1\}$ and the maximal power is 2.\label{fig:examples}}
\end{SCfigure}

For simplicity, we start with polynomial type superpotentials up to a maximal degree:
\begin{equation}
 W=\sum_{n=0}^{N_{\rm max}}\alpha_n \phi^n~.
\end{equation}
The coefficients are complex-valued and its real and imaginary part are initially drawn from a uniform distribution in a given range $(-x,x).$ We then normalise the input such that the maximal absolute value of the real and imaginary part in the interval of choice for the superpotential $(-z,z)$ is $1.$ We report in due course on our choices of parameters when we describe our numerical experiments. In Figure~\ref{fig:examples} we show one example of the associated scalar potential for such a polynomial superpotential which is given as
\begin{equation}
 V=\frac{\partial W}{\partial \phi}\frac{\partial \bar{W}}{\partial \bar{\phi}}~.
 \label{eq:scalarpotential}
\end{equation}

\begin{figure}
\begin{center}
\includegraphics[width=0.4\textwidth]{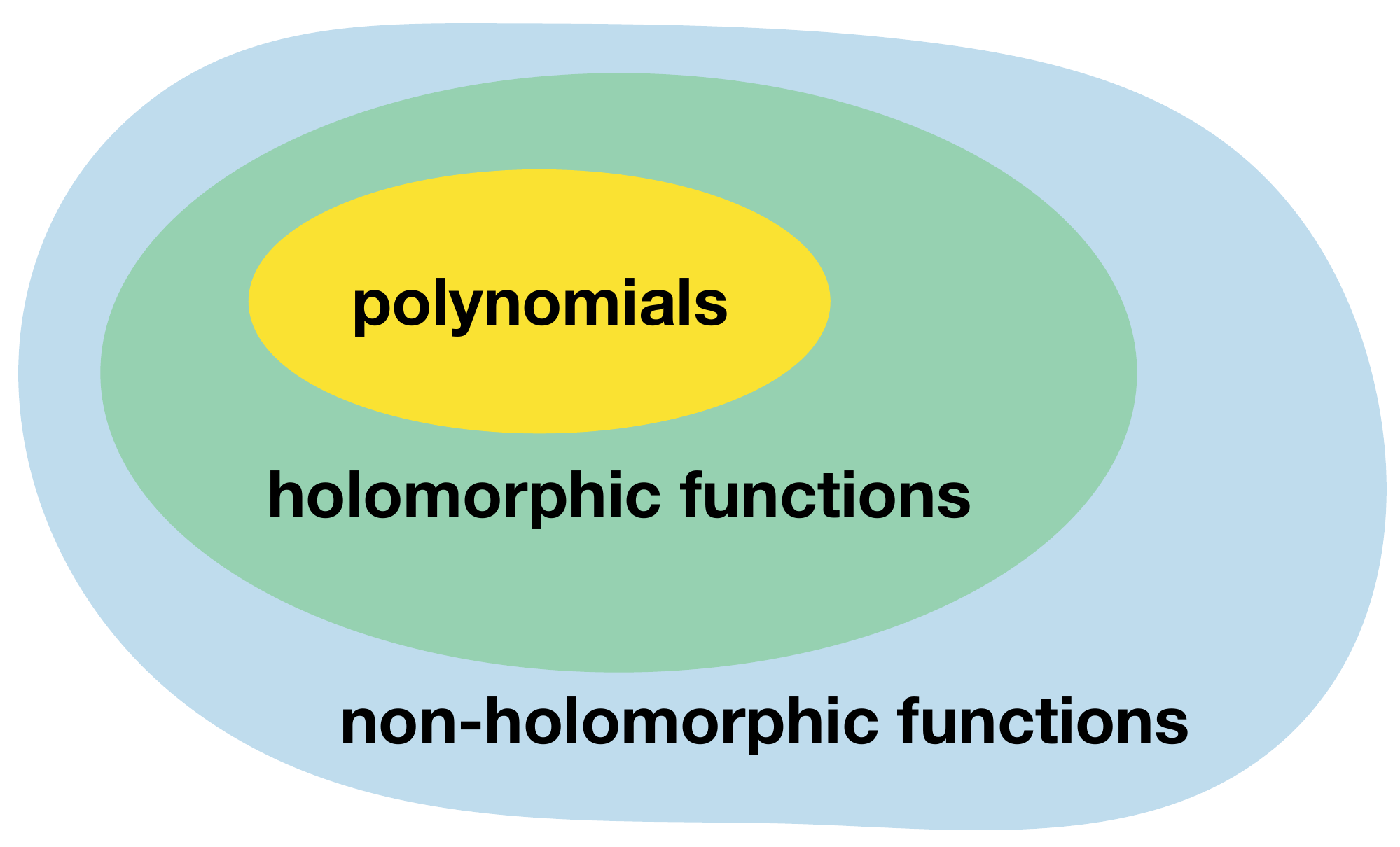}
\end{center}
\vspace{-0.3cm}
\caption{Schematic illustration of the three function spaces which we are dealing with: the space of polynomials (up to a given power), the space of holomorphic functions, and the space of non-holomorphic functions. \label{fig:probabilitydistribution}}
\end{figure}

Our examples are drawn from a probability function which is different to the probability distribution underlying holomorphic functions. The goal of generators is to draw from an underlying probability distribution, typically the same as the input one. In our case, there are three probability distributions which are of interest: 1. The probability distribution of the input superpotentials, which is essentially related to the underlying probability distribution of the polynomial coefficients. 2. The probability distribution associated to general superpotentials. 3. The probability distribution of complex-valued non holomorphic functions.  A cartoon of the three spaces is shown in Figure~\ref{fig:probabilitydistribution}. 

Our goal is to build a discriminating network which only distinguishes between holomorphic and non-holomorphic functions but shall not try to simply explore the `known' polynomial functions. To achieve this, the basic idea is to equip the discriminating network only with the power of checking for the local property (holomorphicity) and not for the global properties required for polynomial checks. Note that this is precisely what certain GAN layouts achieve involuntarily in the context of image generation~\cite{distorted}.

A visualisation of our network layout for this goal is shown in Figure~\ref{fig:overalldesign} in the middle (discriminator) and at the bottom (generator).  The network design is very similar to networks used for generating fake MNIST samples which for instance can be found in~\cite{mnistgansample}. Here we have adopted the structure of the discriminating network to feature convolutional layers with a size of two by two pixels and a stride of one. This is to ensure that the network is capable of checking the local consistency condition of holomorphicity (cf.~\eqref{eq:CRholomorph}). The final activation of the generating network is $\tanh$ to generate a number between $-1$ and $1$ for each image point. The detailed network structure can be found at the end of this article. 

\begin{figure}
\begin{center}
\includegraphics[width=0.5\textwidth]{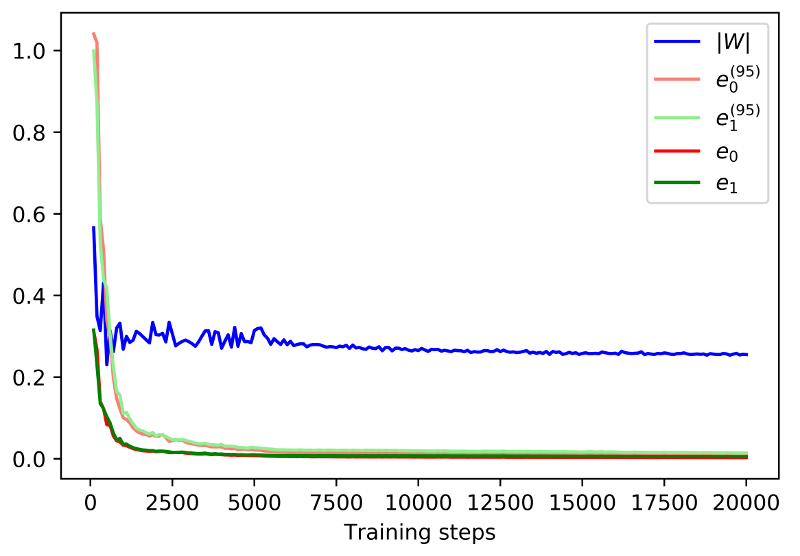}
\end{center}
\vspace{-0.5cm}
\caption{Evolution of the network; comparing the scales of the superpotential and the mean errors as defined in the mean text. \label{fig:evolutionoferror}}
\end{figure}

\begin{SCfigure}
\includegraphics[width=0.7\textwidth]{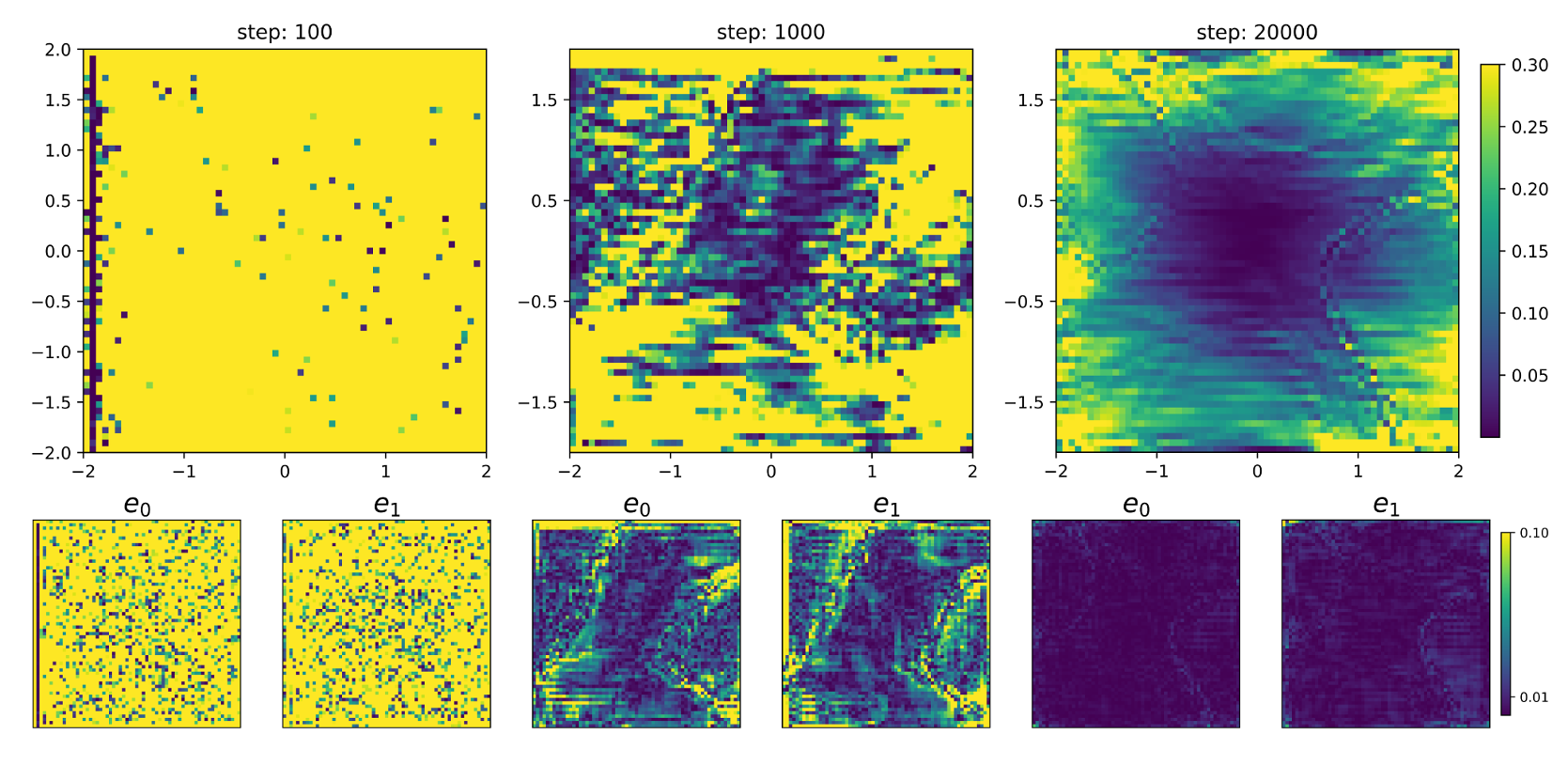}
\caption{Top: Evolution of the potential for a fixed noise input. Snapshots are taken after steps: 100 (beginning), 1000 (middle), 20000 (end). The normalisation of the colour-bar is taken to be the same in all three instances
Middle: Evolution of the error $e_1$ and $e_2$ as defined in Equation~\eqref{eq:error}. The snapshots are taken at the same times as for the potential and the grid is the same as for the potential. Again the colour-bar is the same for all six plots. Videos involving multiple examples will be available online.
\label{fig:snapshots}}
\end{SCfigure}

For our training set we use 10.000 polynomial superpotentials, which are generated from a choice of underlying parameters as described above. We then train our network using the RMSprop optimiser and a batch size $256.$ Our implementation is based on \texttt{tensorflow} and \texttt{Keras}. We have performed hyperparameter tuning regarding the optimiser. We present examples in this letter based on a learning rate of $2e-4$, a decay rate of $6e-8.$ In the following, we present results based on a training set with polynomials of degree $2$, range of coefficients $\{-1,1\},$ and a box size of $\{-2,2\}.$ We have performed tests with polynomials up to degree $5$, varied the ranges of coefficients from $\{-1,1\}$ to $\{-5,5\},$ and box sizes of length $\{2,4,6\}.$ We have also searched over different grid sizes. Before turning to the results, let us briefly comment on how the results from the trained generator have to be scrutinised in two ways:
\begin{enumerate}
 \item Can the numerical solutions be seen as holomorphic functions, given that there are inevitably numerical errors present? To define the error as the deviation from the Cauchy-Riemann equations~\eqref{eq:CRholomorph} is incomplete, as it does not allow the comparison on dimensional grounds to the actual scales involved in the potential. On dimensional grounds, we hence multiply with a length scale $\delta z$, here taken to be the lattice spacing. The errors are then
\begin{eqnarray}
\nonumber e_1&=&\delta z\left(\frac{\partial u}{\partial x}-\frac{\partial v}{\partial y}\right),\\
e_2&=&\delta z\left(\frac{\partial u}{\partial y}+\frac{\partial v}{\partial x}\right). \label{eq:error}
\end{eqnarray}
This error should be small compared to the scales involved in the superpotential. Comparing the error at each point in the grid with the corresponding superpotential value is mis-leading as the superpotential can vanish but its derivatives do not. To avoid this problem, we look at the distribution of errors, its mean and the respective 95 percent confidence level, where the latter is taking into account the spread of the error. We confront these values with the mean absolute value of the superpotentials. For potentials with interesting properties we also perform a visual check whether there seems to be a correlation between the errors and the structure of the potential.
 \item Are the numerical solutions well approximated by polynomial superpotentials? Our aim is to have results which are not necessarily fit by polynomials to explore the space of holomorphic functions. The basic idea is to fit with a polynomial the real data and see that a fit to the generated data is not a good fit. Different methods might be suitable to perform this task, here we use a method based on least square optimisation. To establish whether the generated results are polynomials of a particular degree, we perform a least square fit to a general polynomial of that degree, using the real and imaginary part as separate data points, i.e. minimising:
\begin{equation}
 \sum_i(O_i-E_i(\alpha))^2~,
 \label{eq:sqfit}
\end{equation}
where $O_i$ denotes the discrete data points which have been generated and $E_i$ the corresponding parameters obtained from a model with parameters $\alpha$ \footnote{Note that the standard deviation for each data point is in both cases determined by the negligibly small numerical error. Hence we restrict ourselves on the least square values.}. In the case of the training data, the fit clearly reproduces the original coefficients; in particular giving vanishing coefficients for powers higher than present in the original polynomial. Conversely, the fit worsens when the fitting polynomial is of lower degree than the original polynomial. Applied to the generated data, this method can signal that the generator creates functions in a larger class than the one of the training data.
 
\end{enumerate}
A sample of the evolution of our generating network for fixed noise input is shown in Figure~\ref{fig:snapshots}, where we show the scalar potential and the errors at different training steps. From a completely noisy output, the network is trained to produce outputs which, on visual inspection, look similar to polynomials we have started with (cf.~Figure~\ref{fig:examples}), some with notable differences though.The errors are initially very noisy as expected and are getting significantly smaller as desired. The evolution of the expectation value of the absolute value of the superpotential averaged over the entire grid and over $16$ fixed noise inputs, the mean errors and their respective 95\% confidence value is shown in Figure~\ref{fig:evolutionoferror}. We clearly see that the errors, upon training, are becoming smaller than the superpotential expectation value as desired. Hence the network identifies up to small errors what a holomorphic function is.

\begin{figure}
\begin{center}
\includegraphics[width=0.22\textwidth]{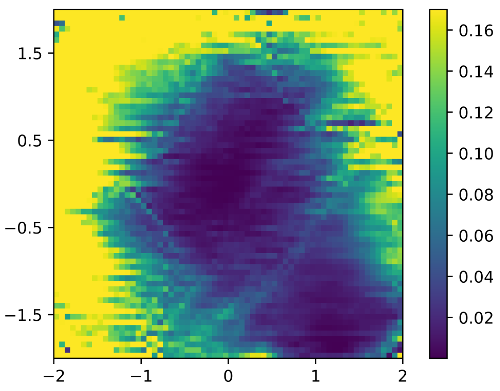}
\includegraphics[width=0.22\textwidth]{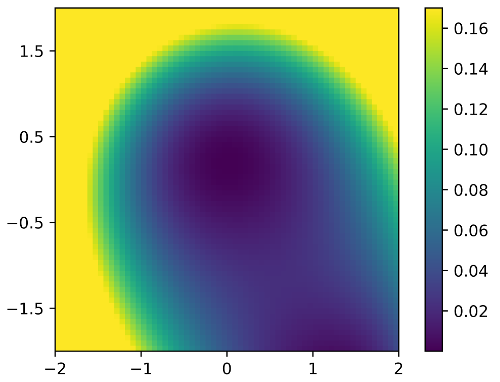}
\end{center}
\caption{Example of a generated potential with two minima from our network which was trained with a set consisting only of quadratic superpotentials. Left: the generated potential. Right: Best fit obtained from a degree $5$ polynomial to the superpotential, clearly showing two minima.\label{fig:generatedexamples}}
\end{figure}

Checking whether the generated superpotentials are of degree two, we find that the generator is producing solutions clearly going beyond polynomials of degree two. A simple check reveals that the initial set of solutions has maximally one minimum, whereas some solutions have multiple minima (cf.~Figure~\ref{fig:generatedexamples}). 
When fitting the generated polynomials to the potentials of varying degree, we note that the fits get better the higher the polynomials are (unlike for the training data). The mean values of the cost~\eqref{eq:sqfit} are
\begin{equation}
 18.02,~5.46,~3.05 ~\text{ for degrees } (2,3,5)~,
\end{equation}
which is an average over $10.000$ generated examples. This generator has been able to identify consistent superpotentials of a type ``unknown'' to it. For degree $2$ polynomials we can clearly visualise the difference, and see that it finds solutions which are clearly physically distinct to solutions in the training set. 

The next step, on which we only comment briefly here, is an analysis on which analytic models the generator has produced. The aim is to find an analytic function which shares the properties of the noisy numerical potential (e.g.~the number of minima and the overall shape of the potential). As an example along these lines, we have performed fits of polynomial models with varying degrees (cf.~Figure~\ref{fig:generatedexamples} for a polynomial fit of degree 5). For complicated models, it would be necessary to fit with other functions (e.g.~exponentials, logarithms, etc.).

Further explorations along these lines are clearly exciting. However, at this stage we leave it for the future and only comment on some of the applications we can envision.\\

\noindent{\bf Outlook:}\\
We can envision a lot of applications and future developments of such generating networks. Let us list a couple of examples:
\begin{itemize}
 \item The class of polynomial potentials clearly provides not the most sophisticated examples we can envision, but shall be seen as a good toy example on establishing the structural difference between the test set and the generated set. Following a similar strategy it will be very intriguing to scrutinise more sophisticated classes of functions.
 \item In the context of supersymmetric model building, along the lines of our work, another application is, which models of supersymmetry breaking can we find. How do our properties generalise to systems with multiple fields? Which properties of the potential in the context of (post-)inflationary cosmology can we obtain?
 \item In the context of string compactifications, can we obtain further consistent compactifications  lying outside of the realm of current models. For instance, can physical systems with particular features (e.g. spectral properties, mass hierarchies) be constructed. 
 \item Such generators are of crucial importance to make distinct statements on mapping out the swampland~\cite{Ooguri:2006in}, i.e.~which directions in theory space are no-go areas, and to generate experimental predictions of string theory. To make such statements precise, we clearly have to know about the solution space of string theory which is out of the reach of current technology apart from small classes of models. 
 \item An interesting avenue will be to build generators which can treat the theory directly on the Lagrangian level. A setup which can generate consistent Lagrangian theories subject to checking symmetry conditions would open several doors. For instance, it would be exciting to explore than supersymmetric solutions which involve non-trivial background gauge fields or non-linear realisations of supersymmetry. Which supergravity solutions could be recovered with such techniques.
\end{itemize}

Overall this paper should be seen as a proof of concept and not as aiming at an extensive analysis. Although clearly interesting, studying the available generator techniques is beyond the scope of this article. We simply demonstrated, using one technique, that this approach can lead to interesting models which go beyond the known models, i.e.~the models known to the network. It will be clearly interesting to build a generator where we can steer the deviation from polynomial equations, i.e.~set how far away it should be. We think that this result is very intriguing as it opens up the possibility to explore new models in fundamental physics with generating techniques. 

We are very much looking forward in going beyond ``line 4'' in the context of particle physics models~\cite{silver2017mastering}. It is exciting to see which model building intuition and sophistication the computer can achieve.

{\it Acknowledgements:} It is a pleasure to thank Ben Hoyle, Fabian Ruehle for discussion. Special thanks to Ivo Sachs for supportive discussions and the initial stimulus which motivated this project. HE is supported by a Carl Friedrich von Siemens Research Fellowship of the Alexander von Humboldt Foundation. SK's research is funded by ERC Advanced Grant ``Strings and Gravity'' (Grant No.~320040).\\
\onecolumngrid

\twocolumngrid
\noindent{\bf Details on neural network architecture:}\\
In Tables~\ref{tab:discriminator} and~\ref{tab:generator} we list the detailed layer structure of our discriminator and generating networks used in this work. During training we have varied the relative training rates of the discriminating and generating networks. We have varied the batchsizes (128,256). For some choices of test sets we observed mode collapse. Again, we stress that the purpose of this article is not to identify the best network design for all possible input configurations. Also we have not yet investigated, although interesting, other box shapes, which correspond to different coverings of the complex plane. We noted that edge effects might play a role, i.e.~the errors tend to be larger at the edges. To avoid such problems, one possibility is to start with a larger grid and then restrict later to the potential on the smaller grid. The generated noise input is drawn from a uniform distribution in the range $(-1,1).$

\begin{table}[t]
\begin{tabular}{l l l l}
 Layer (type)    &   Parameters   &       Output Shape        &      Param. \# \\ \hline \hline   
Input & &(64, 64, 2) &        0        \\ \hline
Conv2D     & (f:8,k:2) &      (64, 64, 8)  &       72       \\ \hline
LeakyReLU & $0.2$ &  (64, 64, 8)   &      0       \\  \hline
Dropout      & 0.4 &   (64, 64, 8)   &      0         \\ \hline
Conv2D       & (16,2) &   (64, 64, 16)  &      528       \\ \hline
LeakyReLU & 0.2 & (64, 64, 16)  &      0         \\ \hline
Dropout      & 0.4 &   (64, 64, 16)  &      0         \\ \hline
Conv2D       & (32,2) &   (64, 64, 32)  &      2080      \\ \hline
LeakyReLU & 0.2 & (64, 64, 32)  &      0        \\ \hline 
Dropout      & 0.4 &   (64, 64, 32)  &      0        \\ \hline 
Conv2D       & (64,2)  &   (64, 64, 64)  &      8256     \\ \hline 
LeakyReLU & 0.2 &  (64, 64, 64) &       0        \\ \hline 
Dropout      & 0.4&   (64, 64, 64)  &      0         \\ \hline
Flatten      & &   (262144)      &      0        \\ \hline 
Dense        &   &   (1)          &       262145    \\ \hline
Activation & sigmoid & (1)          &       0         \\ \hline\hline
Total params:& & &273,081  
\end{tabular}
\caption{Detailed layout of the discriminator network used for our sample.\label{tab:discriminator}}
\end{table}

\begin{table}[t!]
\begin{tabular}{l l l l}
Layer (type)    & Parameters &            Output Shape  &              Param. \# \\ \hline \hline    
InputLayer  &  noise   &    (100)        &         0     \\ \hline    
Dense    &        &    (65536)       &        6619136   \\ \hline
Batch-Normalisation& mom: 0.9 &   (65536)     &          262144    \\ \hline
Activation & ReLU  &  (65536)     &          0         \\ \hline
Reshape      &  &    (16, 16, 256)   &      0         \\ \hline
Dropout       & 0.4 &   (16, 16, 256)   &      0         \\ \hline
UpSampling2D & & (32, 32, 256)  &       0         \\ \hline
Conv2D& (f:128,k:3)  & (32, 32, 128)  &       295040    \\ \hline
Batch-Normalisation & 0.9 &(32, 32, 128)  &       512       \\ \hline
Activation  & ReLU  & (32, 32, 128)  &       0         \\ \hline
UpSampling2D&  & (64, 64, 128) &        0         \\ \hline
Conv2D & (64,3) & (64, 64, 64)  &        73792     \\ \hline
Batch-Normalisation & 0.9 &(64, 64, 64)   &       256       \\ \hline
Activation  & ReLU &   (64, 64, 64)   &       0         \\ \hline
Conv2D& (32,2) & (64, 64, 32)   &       8224      \\ \hline
Batch-Normalisation& 0.9 & (64, 64, 32)    &      128       \\ \hline
Activation  & ReLU &  (64, 64, 32)  &        0         \\ \hline
Conv2D& (2,2) & (64, 64, 2)   &        258       \\ \hline
Activation  & tanh  & (64, 64, 2)    &       0         \\ \hline\hline
Total params: & & &7,259,490
\end{tabular}
\caption{Detailed layout of the generator network used for our sample.\label{tab:generator}}
\end{table}
\bibliography{susygansbibliography}

\providecommand{\href}[2]{#2}\begingroup\raggedright\begin{thebibliography}{10}

\bibitem{silver2017mastering}
D.~Silver, J.~Schrittwieser, K.~Simonyan, I.~Antonoglou, A.~Huang, A.~Guez,
  T.~Hubert, L.~Baker, M.~Lai, A.~Bolton, {\em et.~al.}, {\it Mastering the
  game of go without human knowledge},  {\em Nature} {\bf 550} (2017), no.~7676
  354.

\bibitem{ORaifeartaigh:1975nky}
L.~O'Raifeartaigh, {\it {Spontaneous Symmetry Breaking for Chiral Scalar
  Superfields}},  {\em Nucl. Phys.} {\bf B96} (1975) 331--352.

\bibitem{Intriligator:2006dd}
K.~A. Intriligator, N.~Seiberg, and D.~Shih, {\it {Dynamical SUSY breaking in
  meta-stable vacua}},  {\em JHEP} {\bf 04} (2006) 021,
  [\href{http://xxx.lanl.gov/abs/hep-th/0602239}{{\tt hep-th/0602239}}].

\bibitem{goodfellow}
I.~Goodfellow, J.~Pouget-Abadie, M.~Mirza, B.~Xu, D.~Warde-Farley, S.~Ozair,
  A.~Courville, and Y.~Bengio, {\it Generative adversarial nets},  in {\em
  Advances in Neural Information Processing Systems 27} (Z.~Ghahramani,
  M.~Welling, C.~Cortes, N.~D. Lawrence, and K.~Q. Weinberger, eds.),
  pp.~2672--2680.
\newblock Curran Associates, Inc., 2014.

\bibitem{Albertsson:2018maf}
K.~Albertsson {\em et.~al.}, {\it {Machine Learning in High Energy Physics
  Community White Paper}},  \href{http://xxx.lanl.gov/abs/1807.02876}{{\tt
  1807.02876}}.

\bibitem{darkmachines}
``{Dark Machines (white paper in preparation)}.'' \url{darkmachines.org}, 2018.
\newblock [Online; accessed 17-August-2018].

\bibitem{Paganini:2017hrr}
M.~Paganini, L.~de~Oliveira, and B.~Nachman, {\it {Accelerating Science with
  Generative Adversarial Networks: An Application to 3D Particle Showers in
  Multilayer Calorimeters}},  {\em Phys. Rev. Lett.} {\bf 120} (2018), no.~4
  042003, [\href{http://xxx.lanl.gov/abs/1705.02355}{{\tt 1705.02355}}].

\bibitem{Ravanbakhsh:2016xpe}
S.~Ravanbakhsh, F.~Lanusse, R.~Mandelbaum, J.~Schneider, and B.~Poczos, {\it
  {Enabling Dark Energy Science with Deep Generative Models of Galaxy Images}},
   \href{http://xxx.lanl.gov/abs/1609.05796}{{\tt 1609.05796}}.

\bibitem{distorted}
T.~Salimans, I.~Goodfellow, W.~Zaremba, V.~Cheung, A.~Radford, X.~Chen, and
  X.~Chen, {\it Improved techniques for training gans},  in {\em Advances in
  Neural Information Processing Systems 29} (D.~D. Lee, M.~Sugiyama, U.~V.
  Luxburg, I.~Guyon, and R.~Garnett, eds.), pp.~2234--2242.
\newblock Curran Associates, Inc., 2016.

\bibitem{Aldazabal:2000sa}
G.~Aldazabal, L.~E. Ibanez, F.~Quevedo, and A.~M. Uranga, {\it {D-branes at
  singularities: A Bottom up approach to the string embedding of the standard
  model}},  {\em JHEP} {\bf 08} (2000) 002,
  [\href{http://xxx.lanl.gov/abs/hep-th/0005067}{{\tt hep-th/0005067}}].

\bibitem{mnistgansample}
R.~Atienzia, ``{Advanced Deep Learning with KERAS}.''
  \url{https://github.com/PacktPublishing/Advanced-Deep-Learning-with-Keras},
  2018.
\newblock [Online; accessed 17-August-2018].

\bibitem{Ooguri:2006in}
H.~Ooguri and C.~Vafa, {\it {On the Geometry of the String Landscape and the
  Swampland}},  {\em Nucl. Phys.} {\bf B766} (2007) 21--33,
  [\href{http://xxx.lanl.gov/abs/hep-th/0605264}{{\tt hep-th/0605264}}].

\end{thebibliography}\endgroup
\bibliographystyle{JHEP}

\end{document}